\pdfoutput=1

\documentclass[11pt]{article}

\usepackage[review]{emnlp2021}

\usepackage{multirow}
\usepackage{arydshln}

\usepackage{times}
\usepackage{latexsym}

\usepackage{amssymb}
\usepackage{hyperref}
\usepackage{graphicx}

\newcommand{\done}{\makebox[0pt][l]{$\square$}\raisebox{.15ex}{\hspace{0.1em}$\checkmark$}}
\newcommand{\notdone}{\makebox[0pt][l]{$\square$}{\phantom{---}}}
\usepackage{microtype}
\usepackage{todonotes}
\usepackage[inline]{enumitem}
\usepackage{amsmath}

\usepackage[T1]{fontenc}

\usepackage[utf8]{inputenc}

\usepackage{microtype}

\title{Domain Adaptation for Sparse-Data Settings: What Do We Gain by Not Using Bert?}

\newcommand*{\affaddr}[1]{#1} 

\author{{\centering
Marina Sedinkina ~~
Martin Schmitt ~~ 
Hinrich Sch\"{u}tze{\rm}} \vspace{.15cm}
\\
\affaddr{Center for Information \& Language Processing, LMU Munich} \\
\affaddr{\texttt{sedinkina@cis.uni-muenchen.de}}}

\newcounter{notecounter}
\newcommand{\enotesoff}{\long\gdef\enote##1##2{}}
\newcommand{\enoteson}{\long\gdef\enote##1##2{{
			\stepcounter{notecounter}
			\large\bf
			\hspace{100cm}\arabic{notecounter} $<<<$ ##1: ##2
			$>>>$\hspace{1cm}}}}
\enoteson
\enotesoff

\begin{document}
\maketitle
\begin{abstract}
The practical success of much of NLP depends on the
availability of training data. However, in real-world
scenarios, training data is often scarce, not least because
many application domains are restricted and specific.  In this work, we
compare different methods to handle this problem and provide
guidelines for building NLP applications when there is
\textit{only a small amount of labeled training data available for a specific domain}. 
While transfer learning with pre-trained language models outperforms other methods across tasks,
alternatives do not perform much worse while requiring much less computational effort,
thus significantly reducing monetary and environmental cost.
We examine the performance tradeoffs of several such alternatives,
including models that can be trained up to 175K times faster and do not require a single GPU.
\end{abstract}

\section{Introduction}
The performance of models for core NLP problems heavily
relies on the availability of large amounts of training
data. In real-world scenarios, companies often do not have
enough labeled data for their specific tasks for which the
vocabulary or idioms
(e.g., of social media or medical domains)
can differ a lot from general language use. 
To overcome this issue, most  work
adopts \textit{supervised domain adaptation}: there is a
large labeled training set available in the source domain
and an amount of labeled target data that is insufficient
for training a high-performing model on its own. Other
methods, so-called \textit{unsupervised domain adaptation}
methods, only require labeled data in the source
domain. Both approaches are not directly applicable without
access to labeled data in the source domain.
The focus of this work is therefore to study and evaluate
techniques of domain adaptation for different tasks
when \textit{only a small amount of labeled training data is
available in the target domain}. In this  low-resource
scenario, the data is sparse -- the word types appear in the
data only a few times or even just once (i.e., hapax legomena). 
The sparsity also leads to a high rate of out-of-vocabulary (OOV) words.
These problems negatively affect the performance of a classifier. 
In our work, we look at two NLP applications in sparse-data
scenarios -- sentiment analysis (SA) and Part-Of-Speech
(POS) tagging -- and evaluate how different domain adaptation methods fare on these tasks. 

Several methods have been proposed for domain adaptation
with limited training data in the target domain and no
labeled data in the source domain. Much of this work focuses
on word embeddings \citep{word2vec,glove,fasttext} learned
from large unlabeled corpora. This is one of the promising
applications in transfer learning because the prior
knowledge captured in embeddings can be transferred to
downstream tasks with small labeled data sets.  To capture
domain specific characteristics, both domain-specific and
domain-adapted word embeddings \citep{Sarma2018} have been
proposed. Other ways to include information are tuning
pre-trained off-the-shelf word embeddings and a linear
mapping \citep{Bojanowski}.  Embedding models like
Word2Vec \citep{word2vec}, Glove \citep{glove} and
FastText \citep{fasttext} assign a single vector
representation to a word independent of the
context. However, the meaning of a word changes according to
the context in which it is used. Recently developed
pre-trained language models like BERT \citep{BERT} or
Flair \citep{Flair} address this limitation by allowing to
generate context-dependent representations. They furthermore
allow finetuning all of the parameters using labeled data
from the downstream tasks. For this reason, pre-trained
language models have shown tremendous success in transfer
learning -- they outperform many task-adapted
architectures, achieving state-of-the-art performance on
many sentence-level and token-level tasks. The drawback of
this approach is that these accuracy improvements depend on
the availability of exceptionally large computational
resources. As a result, these models are costly to train,
develop, and deploy. This issue was recently raised in the
NLP community, e.g.,
by \citet{strubell-etal-2019-energy} who argue
that even if the expensive computational resources are
available, model training also incurs a substantial cost to
the environment due to the energy required to power the
hardware. To address these problems, several methods have
been proposed to lower memory consumption and increase the
training speed of BERT models.  Thus, models like
DistilBERT \citep{DistilBERT}, ALBERT \citep{Albert}, and
I-BERT \citep{kim2021ibert} have been introduced. While
these models are lighter, they are still computationally
expensive and require multiple instances of specific
hardware such as GPUs.  This severely limits the access to
pre-trained language models for those who cannot afford
buying or renting such hardware.  In our work, we provide
the answers to two main questions: (i) What do we gain by
not using BERT?  (ii) Which alternative models can be used
to successfully build an application in the sparse-data
scenario of a specific domain?

In our study, we find that the optimal choice of the domain
adaptation method depends on the task:
domain-adapted word embeddings are only beneficial for SA, but for POS tagging, it is better to use
task-adapted (syntactically-oriented) embeddings.
Besides, further finetuning of
word embeddings hurts the performance of sentiment analysis
though it is useful for POS tagging.
Surprisingly, despite
the success of domain-adapted word embeddings in deep
models, the shallow bag-of-words (BoW) approach still outperforms
these methods in sparse-data settings.  
Despite the fact that pre-trained language
embedding models outperform all other domain adaptation methods, they provide a poor speed-accuracy tradeoff for both tasks. For example, for SA the BoW approach is 175K times faster than pre-trained language models while only losing 3 accuray points.
We observe a similar scenario for POS tagging -- finetuning of task-adapted embeddings provides the best speed-accuracy tradeoff. 




\section{Overview of Methods}

\subsection{Categorization of Domain Adaptation Techniques}
In general, a domain consists of a joint distribution over a feature space and a label set \citep{Pan10asurvey}. Differences between domains can be characterized by their marginal distribution over the feature space and label set. 
The training and test data must be from one domain to achieve good performance. The quality of the performance also depends on the amount of training data. Since specific domains often lack sufficient training data, \textit{domain adaptation (DA)} techniques have been proposed to facilitate the transfer of knowledge from one domain (source) to a specific domain (target).
Table \ref{table:settings} shows our categorization of
domain adaptation techniques according to \emph{domain
  adaptation setting} and \emph{resource availability}
(high, low and  zero resource).
\textit{Supervised domain
  adaptation $DA_{sup}$} uses the knowledge from labeled
source domain $D_s$ and applies this knowledge to a target
domain $D_t$, given the labels in both domains
\citep{chelba2006adaptation, daume2009frustratingly,
  pan2010cross,chen2012marginalized}.
\textit{Semi-supervised adaptation} addresses the problem
that the labeled data may be too sparse to build a good
classifier, by making use of a large amount of unlabeled
data (in $D_s$ and/or in $D_t$) and a small amount of
labeled data in $D_t$
\citep{Pan10asurvey}. \textit{Unsupervised domain
  adaptation} refers to training a model on labeled data
from a source domain $D_s$ and applying it on a target
domain, with access to unlabeled data in the target domain
$D_t$
\citep{blitzer2006domain,glorot2011domain,raina_self-taught_2007,ganin2016domain,
  miller-2019-simplified, unsup2, FEMA,
  han-eisenstein-2019-unsupervised}.
Since we are
interested in methods for the case of limited annotated data in the target domain $D_t$, we
consider in this work only the second setting --
semi-supervised domain adaptation. We also assume that we do
not have labeled data in the source domain $D_s$ because
real-world scenarios may lack the training data in the
source domain as well.

\begin{table}[h]
    \centering
    \begin{tabular}{|c|c|c|c|}
    \hline

    \multicolumn{1}{|c}{Domain Adaptation}   & \multicolumn{3}{|c|}{$D_t$ labeled} \\ \cline{2-4}
    
    Setting (resource) &   none & small & large \\\hline\hline
    Supervised  (high)   & \notdone &\notdone &\done\\\hline
    Semi-supervised (low)   & \notdone &\done & \notdone \\\hline
    Unsupervised (zero)    & \done &\notdone & \notdone \\\hline

    \end{tabular}
    \caption{Different settings of domain adaptation (DA) including target domain $D_t$ labeled data.}
    \label{table:settings}
\end{table}


\subsection{Semi-Supervised Domain Adaptation}
To handle the case that domains and distributions used in
training and testing are different, transfer learning
approaches have been proposed. Most of these approaches
apply embeddings, either word embeddings \citep{word2vec,
  glove, fasttext} or pre-trained language models \citep{BERT,
  Flair, Roberta}. 
 The idea is to transfer the latent knowledge present in embeddings to
 downstream tasks with small labeled data sets. Since these representations capture different properties of language, they are crucial for domain adaptation in sparse-data settings.

 The representations that are learned during unsupervised learning on unlabeled data can either be frozen or further optimized during supervised learning on labeled data. We refer to further optimization as finetuning.
In the following, we give an overview of these embedding models, either ``frozen'' or finetuned.

\textbf{Generic Word Embeddings.} The classic solution to solve the domain adaptation problem is finetuning of generic word vectors, i.e., initializing the parameters with word embeddings that are pre-trained on a large corpus of source domain $D_s$ \citep{newman-griffis-zirikly-2018-embedding, Sarma2019}. 
The disadvantage of this approach
is that aggressive finetuning may lead to loss of
information from the original dataset \citep{Bojanowski} and
to overfitting when the training is performed with scarce and noisy data \citep{Astudillo}.

\textbf{Domain-Specific Word Embeddings.}
Domain-specific word embeddings can be trained from scratch on a large amount of unlabeled data from the target domain $D_t$. For some domains, there already exist pre-trained models trained on specific data, e.g., Glove Twitter embeddings\footnote{https://nlp.stanford.edu/projects/glove/} or biomedical word embeddings BioWord2Vec \citep{bioword2vec}. 
Domain-specific word embeddings are usually used as input
features for downstream tasks where embedding weights stay
fixed. To achieve good performance, these embeddings should
have a good quality, e.g., by training them on large corpora
with good coverage of the domain.


\textbf{Cross-Domain Word Embeddings.}
In this approach, word embeddings are learned by incorporating the knowledge from both domains $D_s$ and $D_t$ simultaneously \citep{he-etal-2018-adaptive, yang-etal-2017-simple}. Here, the model tries to differentiate between general and domain-specific terms, by minimizing the distance between the source and the target instances in an embedded feature space.
While this method was successfully applied on a sentiment classification task \citep{he-etal-2018-adaptive,yang-etal-2017-simple} and sparse-data medical NER \citep{newman-griffis-zirikly-2018-embedding}, it heavily relies on the amount of training data available. Thus, this method is not applicable if the amount of unlabeled training data is small.

\textbf{Refined Word Embeddings.} 
Refined word embeddings are word embeddings augmented with additional knowledge from specific domain $D_t$, normally by using lexicons and ontologies. The idea is to improve general word embeddings by using valuable information that is contained in semantic lexicons such as WordNet, FrameNet, or the Paraphrase Database \citep{Faruqui, rothe17autoextend}.
In the clinical domain, the quality of embedding models was
substantially improved by using domain knowledge from
ontologies 
\citep{Augmenting, Ling}. 

\textbf{Domain-Adapted Word Embeddings.}
Domain-adapted word embeddings are pre-trained embeddings adapted for a specific domain. The adaptation requires domain-specific embeddings of good quality. These can be pre-trained specific word embeddings like BioWord2Vec or LSA-based word vectors calculated on domain specific data $D_t$ \citep{Sarma2019}.
Generic and domain-specific word embeddings are then combined using different methods:
\begin{enumerate*}[label=(\alph*)]
\item \textit{Concatenation} of the two word embeddings
allowing the incorporation of information from different domains \citep{roberts-2016-assessing, newman-griffis-zirikly-2018-embedding}. 
  \item \textit{Weighted mean} of the two word vectors where
     word weights are word frequency or inverse document
     frequency, which captures whether the word is common or rare across all phrases \citep{Belousov, Sarma2019}. 

  \item \textit{Kernel Canonical Correlation Analysis
    (KCCA)} of the two word embedding spaces. For example, \citet{Sarma2018} combines pre-trained generic word embeddings and domain specific embeddings learned by applying LSA on the domain specific corpus $D_t$. 

\item \textit{Linear Transformation} methods that use monolingual word vector alignment. 
\citet{Bojanowski} apply this approach for the scenario when the language distribution of the data drastically changes over time.
\citet{newman-griffis-zirikly-2018-embedding} use a similar technique for sparse-data medical NER -- they adapt the multilingual approach proposed in \citep{artetxe-etal-2016-learning}.

\end{enumerate*}

\paragraph{Pre-trained language models (PLMs).}
Recently developed transfer learning algorithms like Flair \citep{Flair} and BERT \citep{BERT} propose to produce word vector representations that dynamically change with respect to the context in which the words appear.
For this reason, they turned out to be an effective
replacement for static word embeddings.
These are finetuning based models, i.e., the embedding weights can be adapted to a given target task without substantial task-adapted architecture modifications.
During pre-training, the model learns a general-purpose representation of inputs, and during finetuning (\textit{adaptation}), the representation is transferred to a new task.
Later, \citet{Roberta}
presented a replication study of BERT
pre-training, \textit{Robustly Optimized BERT Pre-training (RoBERTa)},  with alternative design choices and training strategies that lead to better downstream task performance. They also use more data for pre-training, thereby further improving performance on downstream tasks. While this model leads to significant improvement, operating such large models remains
challenging. For this reason, smaller and lighter models of BERT have been proposed. For example, DistilBERT
 \citep{DistilBERT} or ALBERT \citep{Albert} can be finetuned with good performance on downstream tasks, keeping the flexibility of larger models. 


\section{Case Study}
In this section, we look at two specific NLP applications, Sentiment Analysis (SA) and Part-Of-Speech (POS) tagging in a sparse-data scenario, and evaluate how different domain adaptation methods fare on these tasks. 

For sentiment analysis, we use a benchmark dataset of
\textit{Movie Reviews (MR)} \citep{pang-lee-2005-seeing} and
the \textit{SemEval-2017 Twitter (Twi.)}
dataset.\footnote{http://alt.qcri.org/semeval2017/task4/}
Following the setup of \citet{Sarma2019}, we randomly sample
2500 positive and 2500 negative reviews for experiments with
Movie Reviews. For the Twitter dataset, we randomly choose
6000 tweets with a balanced distribution of the three class labels -- positive, negative, and neutral. The datasets do not have dedicated train/dev/test splits, so we create 80/10/10 splits. 

For POS tagging, we use the \textit{Twitter POS} dataset \citep{Gimpel}, which uses the universal POS tag
set composed of 21 different labels. The dataset contains
1639 annotated tweets for training, 710 tweets for tuning,
and 1201 tweets for testing. As a second dataset, we use a
POS tagged corpus built for the biomedical domain:
\textit{BioNLP POS} dataset
\citep{tateisi-tsujii-2004-part}. It consists of 45
different labels and follows the Penn Treebank POS tagging
scheme.\footnote{http://www.geniaproject.org/genia-corpus/pos-annotation}
The first 1000 sentences
are used 
for training, 500 sentences for development, and 500 sentences for testing.

Table \ref{sparse:statistics} gives a few statistics, demonstrating the difficulty of these tasks. The lexical richness metric shows how many times each word type is used on average (average token occurrence).
A smaller value indicates a more difficult learning problem
as there are fewer chances to learn about a word.
For example, each word type in the MR dataset has, on average, 8 occurrences in the text.
The Twitter SA dataset poses a tougher challenge because average token occurrence is only 5.
The most difficult task is POS tagging for Twitter -- each word is used only 2 times on average.
A large number of low-frequency tokens can substantially harm the performance.
The table also demonstrates that, for each benchmark, a large portion of the word types only occurs once in the whole data set (\textit{hapax legomena}),
indicating a large number of domain-specific technical terms.
For SA, almost half of the training data consists of hapax legomena (47\% for MR dataset and 42\% for Twittter).
For POS tagging, the number of hapax legomena is even higher (e.g., 78\% for Twitter). 
 We also observe a large number of out of vocabulary word tokens (OOVs) in the test split for all data sets.
This is especially an issue for POS tagging in the Twitter dataset -- here 70\% of all test tokens are OOVs. 

\begin{table}[h]
    \centering
\begin{tabular}{|l|r|r|r|}
  \hline
data set &  \multicolumn{1}{c|}{lexical}  & \multicolumn{1}{c|}{hapax}  & \multicolumn{1}{c|}{OOVs}  \\
 & \multicolumn{1}{c|}{richness} & \multicolumn{1}{c|}{legomena}  & \multicolumn{1}{c|}{in test} \\ \hline\hline

    
 SA MR &                       11 &    47.17\% & 4.93\% \\ 
  SA Twi. &                      5 &     42.37\% & 7.27\% \\ 
  POS Bio &                       6 &     50.55\% & 14.98\% \\
  POS Twi. & 	                 2 &    78.36\%   & 28.41\% \\ 

 \hline
    \end{tabular}
    \caption{Dataset statistics. Lexical richness is
    measured as the average token occurrence.
    Hapax legomena only occur once in the whole data set. Out of vocabulary (OOV) word tokens do not occur in the respective train split.}
    \label{sparse:statistics}
\end{table}



\subsection{Sentiment Analysis}
\label{subsec:SA}


\textbf{Bag-of-words (BoW).}
A classic solution to solve SA is
to use a linear classification model on BoW representations.
In our experiments, we apply \textit{logistic regression} from the scikit-learn library\footnote{\url{https://scikit-learn.org/}} with default hyperparameters.
We compare the following representations for a collection of text documents:

\begin{enumerate*}[label=(\alph*)]
    \item \textbf{Sparse Vectors.} We represent documents as sparse vectors of the size of the vocabulary, i.e., we convert a collection of text documents to a matrix of token counts. 

    \item \textbf{Generic Word Embeddings.} Each sentence is expressed as a weighted sum of its constituent word embeddings (Glove). We use raw word counts as weights. Glove embeddings were pre-trained on  Common Crawl (840B tokens). They have  dimensionality  300.
    
    \item \textbf{Domain-adapted KCCA Word Embeddings.}
This method uses as input features domain-adapted (DA) word embeddings, formed by aligning corresponding generic (Glove) and specific word vectors with the nonlinear KCCA approach. Specific word vectors can be created using  $D_t$  by applying Latent Semantic Analysis (LSA) \citep{Sarma2018}. Another possible solution is to use already pre-trained domain-specific embeddings.
For MR, we create 300-dimensional LSA-based word embeddings.
For Twitter, we use pre-trained Twitter Glove embeddings.\footnote{https://github.com/stanfordnlp/GloVe}
Twitter Glove embeddings were pre-trained on a large Twitter corpus (2B tweets, 27B tokens). They have a dimensionality of 100.

\item \textbf{Domain-adapted Word Embeddings: Linear Transformation.}
This approach uses domain-adapted vectors created by combining generic (Glove) and domain-specific vectors using a monolingual word vector alignment technique \citep{Bojanowski}. In our experiments, we use the same domain-specific word embeddings as for the KCCA alignment.
\end{enumerate*}



\textbf{Neural Networks with Generic and Domain-Adapted
  Embeddings.} Instead of using a shallow BoW
approach, we utilize generic or domain-adapted word
embeddings as input features for  Convolutional Neural
Networks (CNN) \citep{kim-2014-convolutional} and
Bi-directional LSTMs (BiLSTM)
\citep{conneau-etal-2017-supervised}. 
We experiment both with and without finetuning of the word vectors during training.

\textbf{Pre-trained language models.}
We train a classifier that takes the
output of the [CLS] token as input.
We 
finetune  all pre-trained parameters.
Our models are
 BERT
\citep{BERT} and its successors, RoBERTa
\citep{Roberta}, DistilBERT \citep{DistilBERT}, and ALBERT \citep{Albert}.
 

\subsection{Part-Of-Speech Tagging}\label{subsec:POS}

\textbf{Window Approach.}  Many different methods  have
been employed for POS tagging with various levels of
performance. One of the best POS classifiers is
the \emph{window approach},
i.e.,
classifiers trained on  windows of text
\citep{Collobert}. In our work, we also use the window
approach, which includes features such as preceding and
following context words (4-grams) and handcrafted
features to deal with unknown words. Each word is thus
represented as a concatenation of the handcrafted features
and one of the following word embeddings:

\begin{enumerate*}[label=(\alph*)]
    \item \textbf{Generic Word Embeddings (Glove).} 
    We use the same Glove embeddings as for SA.
    \item \textbf{Syntactic Word Embeddings (SENNA).} 
    SENNA word embeddings are improved word vectors developed especially for syntax problems \citep{ling-etal-2015-two}. These embeddings have 50 dimensions.

\item \textbf{Domain-adapted Word Embeddings.}
In our experiments, we also apply domain-adapted word embeddings: KCCA \citep{Sarma2018} and linear transformation based \citep{Bojanowski} embeddings. They are obtained using pre-trained Glove embeddings and pre-trained domain specific embeddings. 
For the BioNLP POS tagging task, we use 200-dimensional biomedical word embeddings.\footnote{https://github.com/ncbi-nlp/BioSentVec} They were pre-trained with fastText on PubMed and the clinical notes from MIMIC-III. For the Twitter POS tagging task, we use Glove Twitter embeddings as for SA.
\end{enumerate*}

\textbf{Window Approach with Subspace.}
Following \citet{Astudillo}, we improve the window approach
proposed in \citep{Collobert} by using the Non-Linear Sub-space Embedding (NLSE) model. 
NLSE is implemented
as a simple feed-forward neural network model with one
single hidden layer \citep{Astudillo}. The number of hidden
states is set to 100. Window approaches are implemented using Keras.\footnote{\url{https://keras.io/}}

\textbf{Pre-trained language models.}
 We conduct experiments with the same PLMs as for SA,
and additionally also with BioBERT\footnote{\url{https://github.com/dmis-lab/biobert}} and the Flair model \citep{Flair}.
We use the
FLAIR\footnote{https://github.com/flairNLP/flair} library to perform the experiments. FLAIR
allows combining (``stacking'') different embedding
types. Since the combination of forward and backward Flair
embeddings with Glove embeddings turned out to be the best
for POS tagging \citep{Flair}, we perform this experiment
instead of using only Flair embeddings. To train all the POS
tagging models, we use a BiLSTM architecture on top of a
pre-trained language model and finetune all parameters. 


\section{Results}
\subsection{Sentiment Analysis} 

Table \ref{tab:sent} shows the results of different domain adaptation methods for SA. The performance measure is accuracy. The table is interesting in several ways. Comparing BoW models, we can see that sparse vectors (word counts) show the highest performance on both datasets (87.4\% on MR and 83.6\% on Twitter). The use of generic or domain-adapted word embeddings for a BoW model does not improve the accuracy at all. We also see that the simple BoW method is superior to the finetuning of generic or domain-adapted word embeddings using a biLSTM \citep{conneau-etal-2017-supervised} or a CNN \citep{kim-2014-convolutional}. 
Domain-adapted word embeddings are only beneficial without finetuning -- this approach slightly improves BoW results for MR  (87.8\% vs. 87.4\%) as well as for Twitter  (84.8\% vs. 83.6\%). We also observe that further finetuning of word embeddings often decreases the accuracy. 
 For example, ``frozen'' linearly transformed embeddings perform better than the finetuned version for both datasets in both deep models. 
Since these embeddings are already adapted for a specific domain, there is no need to finetune them unless they are of bad quality. 

\begin{table}[h]
    \centering
    \begin{tabular}{|l|l|l|l|}
    \hline
&    Method  & MR & Twitter \\ \hline\hline
&    BoW (sparse) & \textbf{87.4} & \textbf{83.6}
    \\\cdashline{2-4} 
\multirow{9}{*}{\rotatebox{90}{static embeddings}}
&    BoW (generic Glove) & 81.4 & 62.8\\ 
&    BoW (KCCA) & 78.0 & 60.1  \\
&    BoW (Linear Transf) & 81.2 & 60.5 \\ \cline{2-4} 

&    biLSTM (generic Glove) & 87.2 &  83.1\\
&    CNN (generic Glove)  & 85.2 & 81.5 \\
&    biLSTM (KCCA) & 58.6  & \textbf{84.8} \\
&    CNN (KCCA) & 61.2 &  77.5 \\  
&    biLSTM (Linear Transf)  & 85.8  & 83.1  \\
&    CNN (Linear Transf) & \textbf{87.8} & 83.3 \\ \hline \hline
    
\multirow{6}{*}{\rotatebox{90}{finetuning}}
&    biLSTM (generic Glove) & \textbf{86.6} & 82.6  \\
&    CNN (generic Glove) & 80.2  & 80.0 \\  
&    biLSTM (KCCA) & 69.6 & \textbf{83.6}\\
&    CNN (KCCA) & 73.4 &  83.5  \\ 
&    biLSTM (Linear Transf) & 80.2 & 77.6  \\
&    CNN (Linear Transf) & 82.2  & 82.1  \\ 
    
%
    \hline
    \end{tabular}
    \caption{Sentiment analysis accuracy on  Movie Reviews
      (MR) and Twitter. Best accuracy  for each dataset and block is bolded.}
    \label{tab:sent}
\end{table}

\begin{table}[h]
    \centering
\begin{tabular}{|l|r|r|r|r|}
 \multicolumn{5}{c}{Movie Reviews} \\\hline
 model & $t_{c}$ & $t_{g}$ & acc &  $\text{acc}/t_{g}$ \\\hline
 ALBERT &    124710 & 4343 &  89.2 & 0.0205 \\
    RoBERTa &    154710 & 3362 &  90.2 & 0.0268 \\
 DistilBERT &     75660 & 3293 &  90.2 &  0.0274 \\
       BERT &    152790 & 3674 &  \textbf{91.2} & 0.0248 \\\hline
         CNN (LT) &  145   & 9  &  87.8 & 9.75 \\
 BoW & \textbf{0.73}  & \textbf{0.73} &  87.4 &  \textbf{120} \\ \hline
 
\multicolumn{5}{c}{Twitter} \\\hline
      model & $t_{c}$ & $t_{g}$ & acc & $\text{acc}/t_{g}$ \\ \hline
   ALBERT &     13290 & 3643 &  87.8 & 0.0241 \\
    RoBERTa &     96690 & 4778 &  \textbf{89.5} & 0.0187 \\
 DistilBERT &     48140 & 1618 &  85.0 &  0.0525 \\
       BERT &     94700 & 3030  &  85.5 & 0.0282 \\\hline
       CNN (LT)   &  242 & 16.8  &  83.3 & 4.9 \\ 
BoW & \textbf{0.55} & \textbf{0.55} &  83.6 & \textbf{152} \\ \hline
\end{tabular}
    \caption{Speed and accuracy of pre-trained language models vs.\ the sparse BoW model on Movie Reviews and Twitter for sentiment analysis. The training times $t_c$ (only CPU) and $t_g$ (w/ GPU) are in seconds. $t_c = t_g$ for BoW as it does not need a GPU for fast training.}
    \label{time:SA}
\end{table}

In contrast, KCCA embeddings for MR  perform poorly without
finetuning (58.6\% in biLSTM model and 61.2\% in CNN
model). To obtain these embeddings, we used an LSA-based method for constructing domain-specific word embeddings. A small amount of domain-specific data might have a negative influence on the quality of LSA-based embeddings and thereby also on the quality of domain-adapted word embeddings. This might be a reason for such poor performance of KCCA embeddings. 
However, linearly transformed embeddings achieve much better results than KCCA embeddings despite the use of LSA-based embeddings (e.g., 87.8\% vs. 61.2\% for the CNN).  Hence, we can conclude that the method of \citet{Bojanowski} is superior when there is a small amount of specific domain data. 
Analyzing the results on Twitter, we see that KCCA
embeddings provide better results on this dataset than on MR.
For constructing Twitter domain-adapted embeddings, we used Twitter Glove embeddings, pre-trained on a large unlabeled Twitter dataset. This again demosntrates that the KCCA method is only beneficial when specific embeddings are of better quality.

Table \ref{time:SA} compares PLMs with the best performing neural network model (CNN that uses linearly transformed (LT) embeddings) and surprisingly well-performing sparse BoW model.
We see that PLMs outperform all other approaches.
RoBERTa (89.5\% accuracy on Twitter) and BERT (91.2\% accuracy on MR) have the best performance.
As anticipated, DistilBERT and ALBERT provide comparable results with BERT.
These models were developed to decrease the training time and the number of parameters, keeping the flexibility of larger models.
Besides pure performance, Table \ref{time:SA} also compares the speed-accuracy tradeoff on MR and Twitter datasets.
It shows the CPU and GPU runtime $t_{c/g}$ in seconds.
We see that using DistilBERT and ALBERT models leads to a better speed-accuracy tradeoff for both datasets
if there is no GPU available.
With a GPU this advantage vanishes for ALBERT.
DistilBERT is the clear winner among PLMs -- it achieves $\text{acc}/t_g=0.0274$ on MR and $\text{acc}/t_g=0.0525$ on Twitter.
However, a simple BoW (sparse) method provides a much better speed-accuracy
tradeoff compared to the heavyweight PLMs (120 on MR and 153 on Twitter),
raising the question of how much one is willing to pay for a
few points of accuracy.
While preserving a good accuracy, BoW doesn't require significant computational resources,
allowing to reduce monetary and environmental cost by a large margin.
It is also around 4K-8K times faster to train than the state-of-the-art in terms of accuracy -- even with access to a GPU.
Without a GPU, the speedup even increases to around 200,000 times faster.
The next best performing model in terms of speed-accuracy
tradeoff is a CNN model with linearly transformed
embeddings. It is faster than all other PLM models, but
provides almost the same accuracy as BoW  (87.8\% vs. 87.4\%) while being slower (9.75 vs. 120).  

Taken together, our experimental results favor the simple BoW model with word counts for sparse-data scenarios. 
Despite the simplicity of the method, it achieves  remarkable
performance: (i) It outperforms BiLSTM and CNN methods with domain-adapted word embeddings almost in all cases. (ii) It has a superior speed-accuracy tradeoff when compared to PLMs.





\subsection{Part-Of-Speech Tagging}
Table \ref{tab:pos} compares the performance of different
domain adaptation methods applied for POS tagging. The
measure is
accuracy. Comparing window approaches, it can be seen that the
use of syntactic word embeddings (i.e., SENNA) provides the best
results for the Twitter dataset (86.4\%), and is also
highly beneficial for BioNLP  (92.8\%). Slightly better
results on BioNLP are  achieved using ``Window+Subspace
(Glove)''  (93.1\%). The experiments also show that
the use of domain-adapted word embeddings does not improve
the performance. Both datasets  benefit from
the syntactic SENNA embeddings. Built for tasks involving
syntax, such as part-of-speech tagging or dependency
parsing, these embeddings are sensitive to word order. In a
model where word order is considered, the many syntactic
relations between words are more likely to  be captured properly. This explains the superior performance of SENNA embeddings.

\begin{table}[h]
    \centering
    \begin{tabular}{|l|l|l|l|l|}
    \hline
&              Method &  Twitter & Bio \\ \hline\hline
\multirow{5}{*}{\rotatebox{90}{\footnotesize\begin{tabular}{c}
      static\\ embeddings\end{tabular}}}
&    W (Glove) & 84.5 &  92.2 \\
&    W+Subsp (Glove) &   84.5  &  \textbf{93.1}\\
&    W (SENNA) &  \textbf{86.4} & 92.8 \\
&    W (Linear Transf) &  80.6  & 89.7\\
&    W (KCCA) &  84.0  &  92.7 \\ \hline \hline 
\multirow{7}{*}{\rotatebox{90}{\footnotesize\begin{tabular}{c}
      finetuning \end{tabular}}}
&    W (Glove) &  87.5 &  94.9  \\
&    W+Subsp (Glove) &  88.5   & \textbf{95.2} \\
&    W (SENNA) & \textbf{88.8}  & 94.3\\
&    W (Linear Transf) & 88.5  & 93.9\\
&    W (KCCA) & 87.6  &  94.0 \\ \cline{2-4}    \hline


    \end{tabular}
    \caption{POS tagging accuracy on Twitter and
      BioNLP. W = Window Approach}
    \label{tab:pos}
\end{table}

\begin{table}[h]
    \centering
\begin{tabular}{|l|r|r|r|r|}
 \multicolumn{5}{c}{Twitter} \\\hline
 model & $t_{c}$ & $t_{g}$ & acc &  $\text{acc}/t_{g}$ \\\hline
 ALBERT &    56985 & 7598 &  91.7 & 0.0120 \\
    RoBERTa &    42012 & 5524 &  \textbf{94.9} & 0.0172 \\
 DistilBERT &     22842 & 3013 &  94.1 &  0.0312 \\
       BERT &    41472 & 4597 &  94.8 & 0.0206 \\
        Flair &   2610   & 348  &  92.5 &  0.32 \\\hline 
 W (SENNA) & \textbf{7.5}  & \textbf{7.5} &  88.8 &  \textbf{11.84} \\
 W+Subsp & 83.9 & 83.9 &  88.5 & 1.05 \\ \hline

\multicolumn{5}{c}{BioNLP} \\\hline
      model & $t_{c}$ & $t_{g}$ & acc & $\text{acc}/t_{g}$ \\ \hline
   ALBERT &     55345 & 15813 &  99.8 & 0.0063 \\
    RoBERTa &     37638 & 10912 &  99.8 & 0.0091 \\
 DistilBERT &     20844 & 7149 & \textbf{99.9} &  0.0139 \\
       BERT &     37260 & 8477  &  \textbf{99.9} & 0.0117 \\
       Flair &   1883   & 538  &  99.0 & 0.18 \\
BioBERT & 23520 & 6720 &  98.3 & 0.0146 \\\hline
W (SENNA) & \textbf{5.7} & \textbf{5.7} &  94.3 & \textbf{16.54} \\
W+Subsp & 41.96  & 41.96 &  95.2 &  2.27 \\\hline

\hline
\end{tabular}
    \caption{Speed and accuracy of pre-trained language models vs.\ the sparse BoW model on BioNLP and Twitter for POS tagging. The training times $t_c$ (only CPU) and $t_g$ (w/ GPU) are in seconds. $t_c = t_g$ for W (SENNA) as it does not need a GPU for fast training.}
    \label{time:POS}
\end{table}



For both domains, we see further improvement of the performance when finetuning  pre-trained vectors. For example, finetuning of SENNA vectors allows increasing the accuracy by about 2\% on both Twitter  (88.8\%) and on BioNLP  (94.3\%). We can observe similar behavior in all models -- finetuning of word embeddings improves the results.
Although not all tasks benefit from finetuning the word embeddings further, it helps POS tagging while our experiments on SA showed performance drops for this approach.

Table \ref{time:POS} again shows that PLMs perform
much better than other methods. For example,
the concatenation of Flair and Glove embeddings outperforms all
other window approaches for both Twitter  (92.5\%) and BioNLP
(99.0\%).
A reasonable explanation for this success is that Flair
contextualizes words
based on
their surrounding context and, in addition, models words as sequences of characters.
Since BioBERT was trained on a large amount of data from the biomedical domain, it achieves a very good performance for the BioNLP POS tagging task (98.3\%). However, Table \ref{time:POS}  also shows that BERT (99.9\%), as well as RoBERTa (99.8\%), achieve even better results. 
A compressed version of BERT, DistilBERT, demonstrates results as good as other PLMs -- it achieves an accuracy of 99.9\% for the BioNLP task and an accuracy of 94.2\% for the Twitter task. Analyzing the speed-accuracy tradeoff, we also see that Flair presents a favorable speed-accuracy tradeoff among PLMs 
($acc/t_{g}=0.32$ on Twitter and $acc/t_{g}=0.18$ on BioNLP).

To make good use of a given budget, especially in industrial
settings, one can use an alternative -- the simple window approach with syntactically-oriented SENNA embeddings. This method provides a better speed-accuracy tradeoff compared to the PLMs ($acc/t_{g}=11.84$ on Twitter and $acc/t_{g}=16.54$ on BioNLP).
It is cheap to train and develop, both financially and environmentally. 
The window approach with subspace (W+Subsp) performs worse than W (SENNA) in terms of speed-accuracy (e.g. on BioNLP $acc/t_{g}=2.27$ vs. $acc/t_{g}=16.54$) and only gives an improvement for BioNLP  (94.3\% vs 95.2\%). 

Based on the evidence from these experiments,
PLMs are the best choice for POS tagging in sparse-data
settings if computational resources are unlimited.
However, we have shown that  SENNA embeddings lead to only slightly lower performance while training them is much faster and cheaper.
SENNA word embeddings surpass both generic and domain-adapted embeddings and finetuning them improves the results even further.
Thus, finetuning SENNA embeddings is a reasonable choice to
achieve very good performance and to reduce financial and environmental costs.

\section{Discussion and Conclusion}
The observations from our experiments indicate that 
different tasks should be treated differently, e.g., SA benefits from domain-adapted word embeddings while for POS tagging, it is better to use unadapted task-adapted word embeddings.  
Surprisingly, simple methods (like BoW) outperform BiLSTM and CNN models
with domain-adapted word embeddings. 
Besides, further finetuning of word embeddings often decreases the accuracy for SA while it improves the results for POS tagging in all cases. 
Pre-trained language embedding methods are superior to all
other domain adaptation methods across tasks. However, they
have a very poor speed-accuracy tradeoff. If there are
computational resource constraints (which is typical of both industry and academic research),
domain-specific and sparse-data settings benefit from BoW approaches for SA and from finetuning of task-adapted word embeddings for POS tagging. These methods are extremely fast and efficient, when compared to pre-trained language models, and -- despite their simplicity -- achieve impressive accuracy. 




\bibliography{anthology,custom}
\bibliographystyle{acl_natbib}

\appendix
\section{Appendix}
\subsection{Hyperparameters: Sentiment Analysis. Neural Networks with Generic and Domain-Adapted
  Embeddings.}
 We use the Hedwig\footnote{\url{https://github.com/castorini/hedwig}} library to perform our experiments. For both architectures, we set the batch size to 32, initial learning rate to 0.01 and train the model for 30 epochs. For BiLSTM, we set the number of LSTM
layers to 1. Otherwise, we utilize Hedwig's default parameters.
We experiment both with and without finetuning of the word vectors during training.

\subsection{Hyperparameters: Sentiment Analysis. Pre-trained language models.}

We use the FLAIR
library\footnote{\url{https://github.com/flairNLP/flair}}
for our experiments. We
utilize mini-batch gradient descent. Batch size is 16. We  train
the model for 30 epochs. The initial learning rate is
2e-5. Gradients for backpropagation were estimated using the
Adam optimizer \citep{adam}. Otherwise, we use FLAIR's
default hyperparameters.

\subsection{Hyperparameters: POS Tagging. Pre-trained language models.}
We set the batch size to 32, initial learning rate to 0.1 and train the
model for 150 epochs. The number of hidden states per-layer
of the LSTM is set to 256. Otherwise, we use FLAIR's default hyperparameters.
For the experiments with BioBERT, we use its default parameters.

\end{document}